\documentclass{article}
\usepackage{apalike}
\usepackage{frExamplee}
\usepackage{setspace}

\usepackage{multicol}
\usepackage[bookmarks=true]{hyperref}
\usepackage{xcolor}
\usepackage{amsmath}
\usepackage{mathtools}
\usepackage{algorithm2e}
\usepackage{amsfonts}
\usepackage{multirow}
\usepackage{makecell}
\usepackage{subfig}

\newcommand{\new}[1]{#1}

\begin{document}

\title{Radar-Based Localization For Autonomous Ground Vehicles In Suburban Neighborhoods}

\author{
Andrew J.\ Kramer \\
Amazon, LLC\\
\texttt{1988kramer@gmail.com} \\
\And
Christoffer Heckman \\
University of Colorado, Boulder \\
\texttt{christoffer.heckman@colorado.edu}
}

\maketitle

\begin{abstract}
For autonomous ground vehicles (AGVs) deployed in suburban neighborhoods and other human-centric environments the problem of localization remains a fundamental challenge. There are well established methods for localization with GPS, lidar, and cameras. But even in ideal conditions these have limitations. GPS is not always available and is often not accurate enough on its own, visual methods have difficulty coping with appearance changes due to weather and other factors, and lidar methods are prone to defective solutions due to ambiguous scene geometry. Radar on the other hand is not highly susceptible to these problems, owing in part to its longer range. Further, radar is also robust to challenging conditions that interfere with vision and lidar including fog, smoke, rain, and darkness. We present a radar-based localization system that includes a novel method for highly-accurate radar odometry for smooth, high-frequency relative pose estimation and a novel method for radar-based place recognition and relocalization. We present experiments demonstrating our methods' accuracy and reliability, which are comparable with \new{other methods' published results for radar localization and we find outperform a similar method as ours applied to lidar measurements}. Further, we show our methods are lightweight enough to run on common low-power embedded hardware with ample headroom for other autonomy functions.
\end{abstract}

\section{Introduction}

Deploying autonomous ground vehicles (AGVs) in human-centric environments like sidewalks and multi-use paths remains a significant challenge. High accuracy localization is essential because the difference between the robot driving safely on the sidewalk and drifting into traffic is often just tens of centimeters. Unfortunately, GPS does not meet this accuracy requirement when it is available, and of course one can't assume GPS will be available at all times. Many methods have been published addressing the problem of global localization using cameras and lidar. However, even in ideal conditions these sensors have their weaknesses. Visual odometry is susceptible to low light and motion blur while visual place recognition still has significant problems with appearance changes due to weather, lighting changes, and seasonal changes. Lidar-based methods, on the other hand, are prone to defective solutions when presented with ambiguous scene geometry. These problems may be dismissed as edge cases, but even in a modest-sized fleet of AGVs these edge cases occur quite frequently.

Millimeter-wave radar has the potential to sidestep many of the issues encountered other localization methods. It is not affected by visual challenges like darkness, fog, or smoke. Further, radar is not affected by ambiguous scene geometry because it is able to directly measure motion and sense material properties. Despite these advantages, though, radar has seen limited adoption in the robotic localization literature. Many of the existing methods use the NavTech scanning radar sensor including \cite{cen_2018,cen_2019,barnes_2020}. \cite{cen_2018,cen_2019} focus on the odometry problem, \cite{hong_2020} proposes a SLAM method, and \cite{barnes_2020} adds place recognition and relocalization. Similar works focusing on scanning radars for place recognition \cite{kim2020mulran,kim2021scan} have shown remarkable performance, approaching that of lidar-based place recognition. Unfortunately, the NavTech sensor all of these methods depend on is large, power-hungry, and expensive. 

Automotive-grade millimeter-wave radar sensors are small, low-power, and inexpensive, making them a better option for AGVs. \cite{schuster_2016} uses automotive radar sensors to create maps by using stream clustering to identify landmarks in radar scans. They can then reliably align radar scans to those maps for global localization. \cite{schuster_2016b} aggregates radar detections into pseudo-images and uses the BASD feature detector \cite{rapp_2016} to detect and match features in those pseudo images. \cite{kramer_2020} fuses radar and IMU measurements to estimate a vehicle's velocity and orientation in 3D, and \cite{kramer_2021} extends the method by adding a radar-based occupancy mapping method. The method of \cite{kramer_2021} is similar to \cite{schuster_2016b} in that they aggregate radar detections into discretized maps, but \cite{kramer_2021} uses a learned noise filter based on PointNet \cite{qi_2016} to remove spurious radar detections and they use a 3D octomap \cite{hornung2013octomap} rather than 2D dense pseudo images. However, \cite{kramer_2021} only uses their radar-based map for obstacle detection and does not attempt to localize against it. \cite{doer_2020} and \cite{doer_ekf_2020} present a similar radar-inertial odometry (RIO) method using a Doppler-based radar constraint. A fault with all of these Doppler-based methods is they only use gyroscope measurements to constrain the robot's heading estimate, so their heading estimate drifts severely over longer trajectories. \cite{doer_2021} propose to get around this issue by making the restrictive assumption that the robot is operating in a Manhattan (grid) world. Other techniques \cite{kung2021normal,lu2020milliego} have demonstrated impressive performance, but the former does not include an IMU which would likely improve performance, while the latter treats the IMU with a neural network. \new{There are many ongoing research directions related to the use of automotive radar being developed continuously \cite{harlow2023new}, some of which also include loop closure detection with federated back-ends \cite{10160670} and are resistant to dynamic obstacles and map elements \cite{10207713}.} To the author's knowledge, so far no method has solved the problem of a general heading constraint in RIO using low-cost automotive sensors.

Our contributions beyond these works are as follows:
\begin{enumerate}
    \item A method for RIO that includes a novel method for dynamic outlier rejection and a unique radar-based heading constraint.
    \item A method for creating radar-based maps that effectively removes noise and preserves repeatable environmental features.
    \item A method for estimating the robot's global pose by registering local-to-global radar maps with accuracy and reliability comparable to lidar-based methods.
\end{enumerate}

\section{Radar-Inertial Odometry}
\label{sec:rio}

\begin{figure}[h!]
\centering
\includegraphics[width=\linewidth]{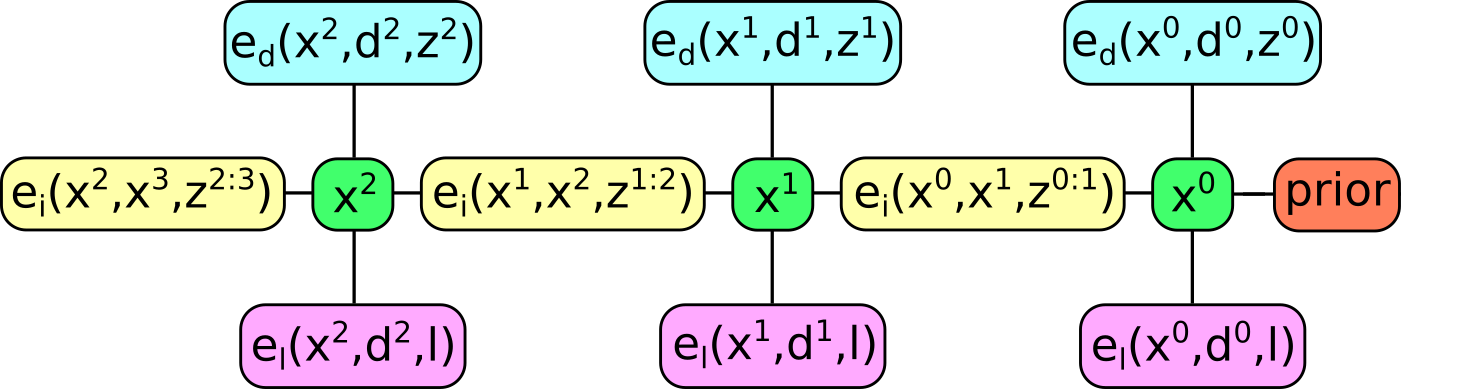}
\caption{Factor graph used for the RIO estimator, with increasing time going left. \new{The Doppler and Landmark errors $e_d$ and $e_l$ provide constraints on the state that results in a nonlinear minimization problem over these errors by adjusting the state variables $x^i$. The states are connected through binary constraints $e_i$ induced by IMU measurements. A prior is formed from marginalized states prior to $x^0$ which have fallen off of the sliding window, thereby maintaining a constant-time algorithm.}}
\label{fig:factor_graph}
\end{figure}

Our RIO method takes inputs from an arbitrary array of synchronized radar sensors and a single IMU. It estimates the robot's state in 6DoF using a nonlinear optimization over a sliding window of recent states and measurements implemented with iSAM2 \cite{kaess_2011}. The overall structure of the problem is shown as a factor graph in Figure \ref{fig:factor_graph}. In subsequent sections we give further detail on the constraints used in the optimization problem including our dynamic object detection method and our Doppler, IMU, and heading constraints. However, we'll first give a brief overview of the notation used in subsequent sections and defined the estimated states and sensor measurements.

\subsection{Notation, State, and Measurements}

We note scalars in lower case as in $r$, vectors in lower case bold as in $\mathbf{v}$, and matrices in upper case bold as in $\mathbf{R}$. The frame in which a quantity is expressed is noted as a pre-superscript. For instance, $\prescript{I}{}{\mathbf{p}}$ denotes vector $\mathbf{p}$ expressed in frame $I$. For velocities, the pre-superscript notes both the frame for \new{and in which velocity is being expressed; note that velocities are always relative to a global reference frame. As an example, $\prescript{R}{}{\mathbf{v}}$ is the velocity of frame $R$ relative to the global reference frame expressed in frame $R$.} We represent rotations as either 3D rotation matrices $\mathbf{R}$ or quaternions $\mathbf{q}$. Rotation frames are noted as pre-scripts. So $\prescript{I}{R}{\mathbf{q}}$ is the quaternion describing the rotation from frame $R$ to frame $I$. The frames used in this work include $O$, the global reference frame; $I$, the IMU reference frame; and $R$ the frame of a radar sensor. Lastly, estimated quanities are noted with a $\hat{}$.

The proposed system optimizes a sliding window of $N$ robot states $\mathbf{X} = [\mathbf{x}^0, \dots, \mathbf{x}^N]$ where the state at timestep $k$ is $\mathbf{x}^k = [\prescript{O}{}{\mathbf{v}}^T,\prescript{O}{I}{\mathbf{q}}^T,\mathbf{b}_a^T,\mathbf{b}_g^T]^T$, the robot's velocity, orientation, and IMU biases. The measurement at a single timestep from a radar sensor is referred to as a scan, and a scan consists of a set of detections $\mathcal{D}$. Each detection represents a target detected by the CFAR \cite{cfar} detector running onboard the radar sensor and detections are parameterized as $\mathbf{d}=[\prescript{R}{}{\mathbf{p}}^T, \prescript{R}{}{\dot{r}}]^T$, \new{where $\mathbf{p} \in \mathbb{R}^3$ refers to the 3-dimensional vector between the radar sensor and the detection, and $\dot{r} \in \mathbb{R}$ refers to the Doppler measurement along the vector (i.e., the velocity of the detection along the ray cast between the sensor and the target).} A measurement from the IMU is parameterized as $\mathbf{z}=[\prescript{I}{}{\mathbf{a}}^T, \prescript{I}{}{\boldsymbol{\omega}}^T]^T$ where $\prescript{I}{}{\mathbf{a}}$ and $\prescript{I}{}{\boldsymbol{\omega}}$ are the acceleration and angular velocity in the IMU frame respectively.

\subsection{Dynamic Object Rejection}

A crucial part of our RIO system is our method for detecting and discarding radar detections which represent dynamic objects like cars, cyclists, and pedestrians. A common way to do this across many sensor types is random sample consensus (RANSAC) \cite{fischler_1981}. There is a simple linear relationship between a static detection's range rate to the sensor's velocity, $\prescript{R}{}{\dot{r}}=\prescript{R}{}{\mathbf{v}}\cdot\frac{\prescript{R}{}{\mathbf{p}}}{\|\prescript{R}{}{\mathbf{p}}\|}$ and it is easily incorporated into a RANSAC model to solve the problem:

\begin{equation}
   \prescript{R}{}{\hat{\mathbf{v}}} = \underset{\prescript{R}{}{\mathbf{v}}}{\text{argmin}}\Bigg(\sum_{\mathbf{d}\in\mathcal{D}}\prescript{R}{}{\dot{r}}_\mathbf{d} - \prescript{R}{}{\mathbf{v}} \cdot \frac{\prescript{R}{}{\textbf{p}}_\mathbf{d}}{\|\prescript{R}{}{\textbf{p}}_\mathbf{d}\|}\Bigg)^2
\end{equation}

The above formulation has been used in other radar odometry methods including \cite{kramer_2020}. It can reliably differentiate between static and dynamic objects in a scan from a single sensor and provide a good initial guess at that sensor's velocity. Using this model with multiple synchronized sensors would require running a separate RANSAC problem for each sensor, but this presents a problem. If a scan from one sensor does not contain a sufficient number of inliers then that scan must be discarded entirely, even if there is a sufficient number of inliers from all the sensors combined. Further, this method results in different estimates of the local frame velocities for each sensor $\prescript{R_s}{}{\mathbf{v}}$ but it would be more useful to estimate a single IMU frame velocity $\prescript{I}{}{\mathbf{v}}$. 

To address these issues we consider all detections from each sensor in the same RANSAC problem. To do this we first transform the detection's location in the radar frame to the IMU frame as $\prescript{I}{}{\mathbf{p}} = \prescript{I}{R_s}{\mathbf{R}}\prescript{R_s}{}{\mathbf{p}} + \prescript{I}{R_s}{\mathbf{t}}$ where $\prescript{I}{R_s}{\mathbf{R}}$ is the rotation from the frame of radar sensor $s$ to the IMU frame and $\prescript{I}{R_s}{\mathbf{t}}$ is the translation. We then offset the range rate measurement with the velocity induced by the lever-arm between the radar sensor and IMU as

\begin{equation}
    \prescript{I}{}{\dot{r}} = \prescript{R_s}{}{\dot{r}} + \Big(\big(\prescript{R_s}{I}{\mathbf{t}}\big)^{\times} \prescript{R_s}{I}{\mathbf{R}}\prescript{I}{}{\boldsymbol{\omega}}\Big) \cdot \frac{\prescript{R_s}{}{\mathbf{p}}}{\|\prescript{R_s}{}{\mathbf{p}}\|}
\end{equation}

\noindent where $(\cdot)^\times$ returns the skew-symmetric matrix corresponding to a given 3D vector. This allows us to process all of the detections from each radar sensor in the same RANSAC problem as

\begin{equation}
    \prescript{I}{}{\hat{\mathbf{v}}} = \underset{\prescript{I}{}{\mathbf{v}}}{\text{argmin}}\Bigg(\sum_{s\in\mathcal{S}}\sum_{\mathbf{d}\in\mathcal{D}_s}\prescript{I}{}{\dot{r}}_\mathbf{d} - \prescript{I}{}{\mathbf{v}} \cdot \frac{\prescript{I}{}{\mathbf{p}}_\mathbf{d}}{\|\prescript{I}{}{\mathbf{p}}_\mathbf{d}\|}\Bigg)^2
\end{equation}

\noindent where $\mathcal{S}$ is the set of radar scans and $\mathcal{D}_s$ is the set of detections from radar sensor $s$. Outliers identified in solving this RANSAC problem are assumed to represent dynamic objects and are discarded.

\subsection{Doppler Constraint}

After discarding detections identified as dynamic in the RANSAC step, radar detections are used to create constraints on the odom frame velocity $\prescript{O}{}{\mathbf{v}}$. The Doppler error function is 

\begin{equation}
\begin{split}
     &e_d(\hat{\mathbf{x}}^k, \mathbf{d}, \mathbf{z}^k) = \\ \prescript{R}{}{\dot{r}} - & \Big(\prescript{R}{O}{\hat{\mathbf{R}}}^k\prescript{O}{}{\hat{\mathbf{v}}}^k + 
     \big(\prescript{R}{I}{\mathbf{t}}\big)^\times \prescript{R}{I}{\mathbf{R}}\big(\prescript{I}{}{\boldsymbol{\omega}}^k-\hat{\mathbf{b}}_g^k\big) \Big) \cdot \frac{\prescript{R}{}{\mathbf{p}}}{\|\prescript{R}{}{\mathbf{p}}\|} 
\end{split}
\end{equation}

\noindent where $\prescript{R}{O}{\hat{\mathbf{R}}}^k=\prescript{R}{I}{\mathbf{R}}\prescript{I}{O}{\hat{\mathbf{R}}}^k$ and $\prescript{I}{O}{\hat{\mathbf{R}}}^k$ is the global frame to IMU frame rotation, which is estimated as part of the state vector; and $\prescript{R}{I}{\mathbf{R}}$ is the IMU frame to radar frame rotation, which is constant. An IMU measurement interpolated to match the timestamp of the radar measurement $\mathbf{z}^k$ is also required to estimate the effect of the sensor's angular rate on the Doppler measurement.

\subsection{IMU Constraint}
\label{sec:imu-constraint}

In our system, IMU measurements are used to measure the system's change in global-frame velocity and orientation between radar measurements. Our IMU model is very similar to that provided in GTSAM \cite{forster_2015}, except we do not use the IMU to measure the change in the system's full pose, only its velocity and orientation. 

The IMU provides measurements at many times the rate of the radar sensors and the IMU and radar measurements are not synchronized. To address this we first interpolate between the IMU measurements immediately before and after the radar measurements to obtain estimated IMU readings that align temporally with the radar measurements at timesteps $k$ and $k+1$. We then use the method for on-manifold IMU pre-integration presented in \cite{forster_2015} to combine the set of IMU measurements between times $k$ and $k+1$ into a single relative constraint which is used to predict the state $\mathbf{x}^{k+1}$ from $\hat{\mathbf{x}}^k$. The IMU error is then defined as the difference between the estimated state $\hat{\mathbf{x}}^{k+1}$ and the IMU propagated state

\begin{equation}
    \mathbf{e}_i(\hat{\mathbf{x}}^k,\hat{\mathbf{x}}^{k+1},\mathbf{z}^{k:k+1}) = \begin{bmatrix}
    2\Big[\prescript{O}{I}{\mathbf{\hat{q}}}^{k+1}\otimes (\prescript{O}{I}{\mathbf{q}}^{k+1})^{-1}\Big]_{1:3} \\
    \prescript{O}{}{\mathbf{\hat{v}}}^{k+1} - \prescript{O}{}{\mathbf{v}}^{k+1} \\
    \mathbf{\hat{b}}_g - \mathbf{b}_g \\
    \mathbf{\hat{b}}_a - \mathbf{b}_a
    \end{bmatrix}
\end{equation}

\noindent where the $\otimes$ operator is as defined in \cite{barfoot_2011}. 

\subsection{Heading Constraint}
\label{sec:heading_constraint}

With only the Doppler and IMU constraints our state estimation problem is underdetermined. While the gyroscope biases in pitch and roll are constrained by the gravity direction, there is no constraint on the gyroscope bias in yaw. This will cause the system's heading estimate to drift over longer trajectories. To address this we implement a novel means of constraining the system's heading estimate by identifying and tracking persistent, repeatable landmarks in radar scans. This consists of two components: a unique method for associating new radar detections to existing landmarks, and a series of checks to determine if the confidence in a given landmark is high enough to be used to create constraints in the estimator. 

To identify candidate matches between the set of new radar detections $\mathcal{D}$ and tracked landmarks from the last timestep $\mathcal{L}$ we first construct a dense matrix of the distances $\mathbf{S} \in \mathbb{R}^{|\mathcal{D}|\times|\mathcal{L}|}$ between each detection $\mathbf{d}$ and landmark $\mathbf{l}$. We use a unique weighted polar distance metric first proposed in \cite{minguez_2006}:

\begin{equation}
\label{eq:distance}
\begin{split}
    \phi(\mathbf{p}) =& \text{ atan2}(\mathbf{p}^y,\mathbf{p}^x) \\
    \mathbf{S}_{\mathbf{d},\mathbf{l}} =& \Big(L^2\big(\phi(\prescript{I}{}{\mathbf{p}}_\mathbf{d}) - \phi(\prescript{I}{}{\mathbf{p}}_\mathbf{l})\big)^2 \\
    &\ + \big(\|\prescript{I}{}{\mathbf{p}_\mathbf{d}}\| - \|\prescript{I}{}{\mathbf{p}_\mathbf{l}}\|\big)^2\Big)^\frac{1}{2} 
\end{split}
\end{equation}

\noindent This distance metric is calculated in polar coordinates and it allows us to control the relative importance of the azimuth and range dimensions by setting the coefficient $L$. This allows us to more accurately model the noise distribution of our radar sensors when matching new detections to existing landmarks. 

\new{This is the only appearance of the position of the Doppler targets $\mathbf{p_d}$ in our algorithm. This is because typical frame-to-frame tracking of these targets is generally not reliable over long durations due to the nature of noise affecting radar measurements. Landmarks are generally not consistently trackable for long traces in time, however they \emph{are} very consistent through rotations. In fact, through rotation in yaw, radar targets are remarkably consistent, likely as a result of the radar cross section of the target not undergoing significant change compared with other types of motion with respect to hypothetical targets. We leverage this fact to support localization through rotation in yaw specifically.}

At this point $\mathbf{S}$ can be viewed as a weighted bipartite graph and a set of potential matches between detections and landmarks is found with the Hungarian algorithm \cite{munkres_1957}. A potential match is validated if the distance between the detection and landmark found by equation \ref{eq:distance} is below a user-defined threshold. New detections for which no valid match is found are initialized as new landmarks. Tracked landmarks are discarded if the time elapsed since the last time the landmark was successfully matched to a new detection is above a threshold value.

Each valid match between a landmark and detection is not necessarily used to create a new constraint in the odometry estimator. We track two quantities for each landmark to determine if a new match should be used to create a new constraint. The number of times the landmark was previously observed must be above a threshold value and the maximum error in the obervations of the landmark must be below a threshold value. If the matched landmark meets these criteria the matched pair $[\mathbf{d},\mathbf{l}]$ is added to the set of active matches $\mathcal{S}$. For each active match an error term of the form:

\begin{equation}
    e_l(\hat{\mathbf{x}}^k,\mathbf{d},\mathbf{l}) = \phi(\prescript{I}{R}{\mathbf{R}}\prescript{R}{}{\mathbf{p}}_\mathbf{d}+\prescript{I}{R}{\mathbf{t}})-\phi(\prescript{I}{O}{\hat{\mathbf{R}}}^k\prescript{O}{}{\mathbf{p}}_\mathbf{l}+\prescript{I}{O}{\mathbf{t}})
\end{equation}

\noindent is added to the estimator. Note the landmark position $\prescript{O}{}{\mathbf{p}}_\mathbf{l}$ and the IMU frame to global frame translation $\prescript{O}{I}{\mathbf{t}}$ are not estimated, they are treated as constants. $\prescript{O}{}{\mathbf{p}}_\mathbf{l}$ is set to the first observed position of the landmark and $\prescript{O}{I}{\mathbf{t}}$ is found by integrating the estimated velocities $\prescript{O}{}{\hat{\mathbf{v}}}$. This greatly simplifies the optimization and ensures the landmark error terms only influence the system's orientation estimate $\prescript{O}{I}{\hat{\mathbf{q}}}$. Empirical testing showed estimating $\prescript{O}{}{\mathbf{p}}_\mathbf{l}$ and $\prescript{O}{I}{\mathbf{t}}$ as part of the system's state actually has negative effects on the accuracy of the system's odometry estimate. \new{Since this method is inevitably paired with a technique that corresponds current radar measurements with a prior map, long-term longitudinal tracking introduces error that creates tension with the prior map tracking, and is therefore not utilized.}

\section{Radar Mapping and Matching}

Our radar-inertial odometry method provides high-frequency, smooth relative motion estimates but does not include a method to constrain drift or localize the robot in a global frame. In the following sections we describe our method for solving these problems. This consists of two components: a method for constructing maps from radar detections described in Section \ref{sec:radar_mapping} and a method for estimating the robot's pose by registering local radar maps to global radar maps described in Section \ref{sec:map_matching}.

\begin{figure}[t!]
\centering
\subfloat[Robot's path through a neighborhood.]{\includegraphics[height = 2.25in]{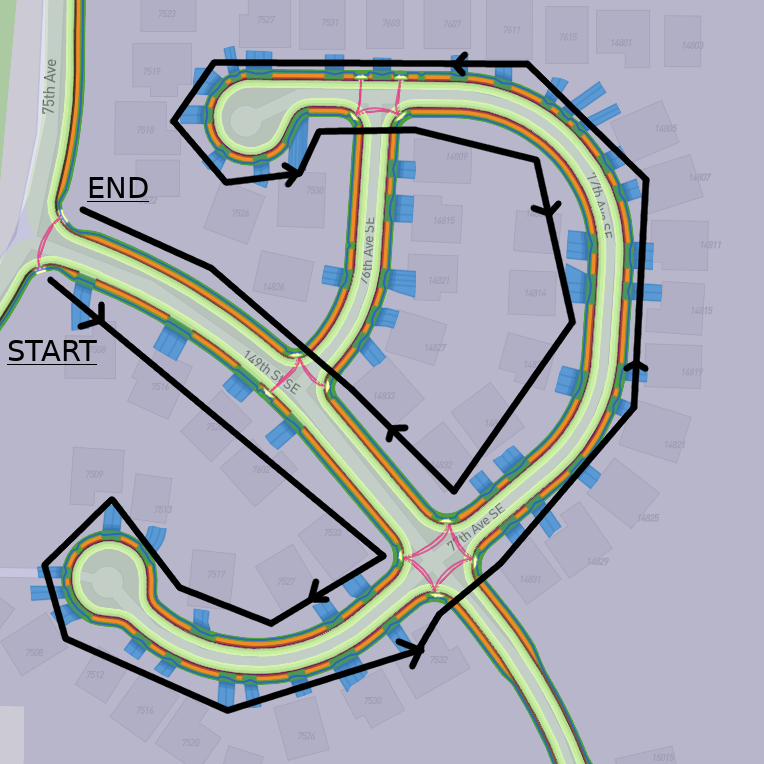} \label{fig:trajectory}}
\subfloat[Radar map generated from that mission.]{\includegraphics[height = 2.25in]{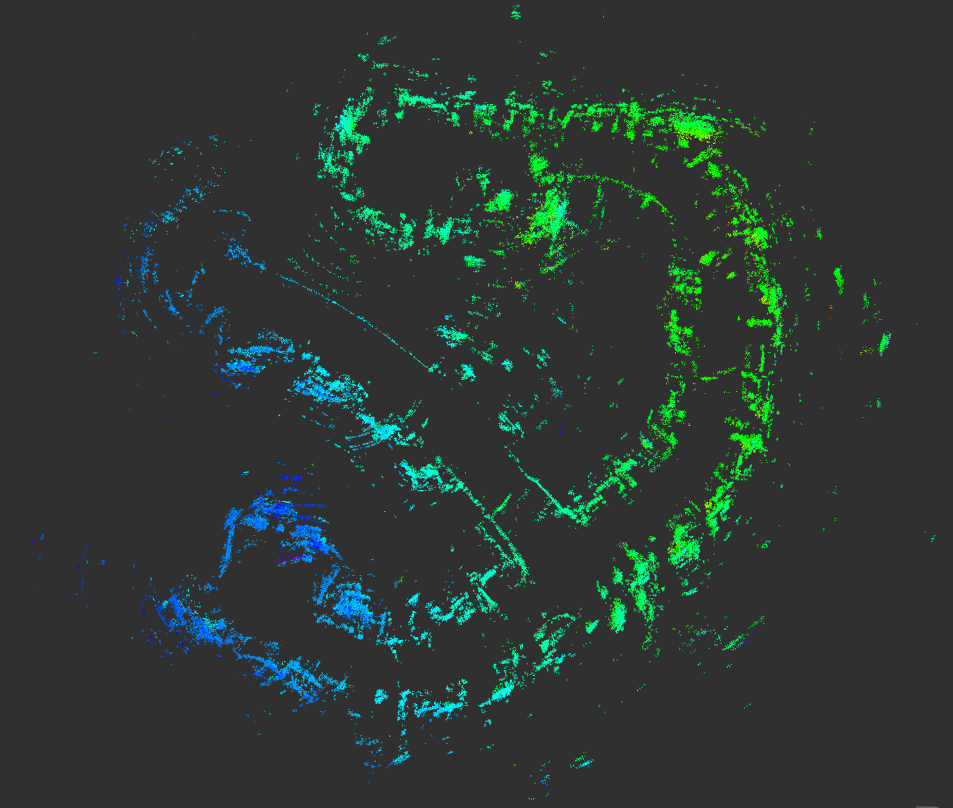} \label{fig:maps}}
\caption{A neighborhood trajectory and single global map generated using the volumetric mapping method.}
\end{figure}

\subsection{Radar Mapping}
\label{sec:radar_mapping}

\begin{figure}[t]
\centering
\subfloat[Raw radar detections from a short mission transformed into a global reference frame.]{%
  \includegraphics[height=2.45in]{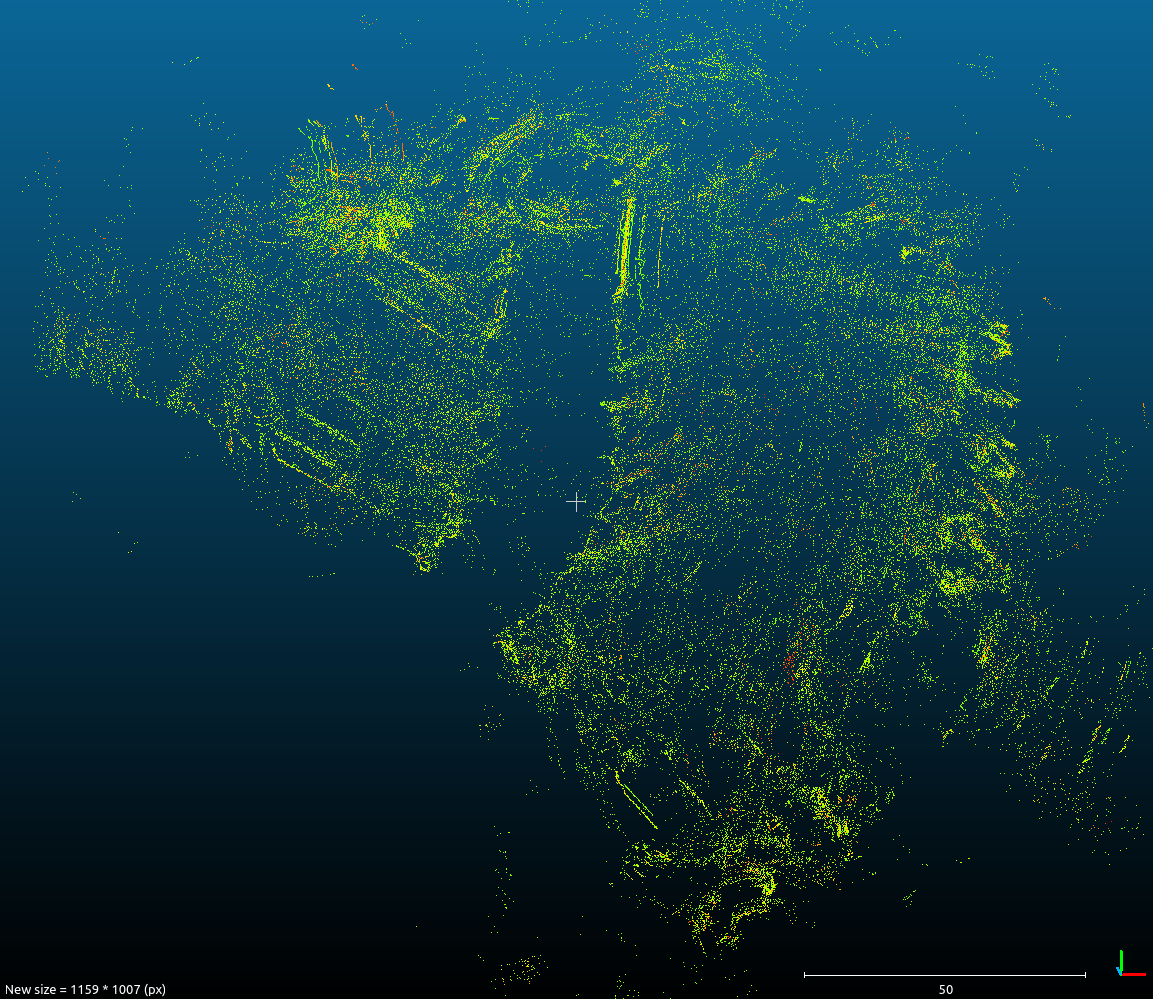}%
  \label{fig:raw_detections}%
}\qquad
\subfloat[The detections in Figure \ref{fig:raw_detections} binned into a spatial histogram colored by the number of detections per voxel.]{%
  \includegraphics[height=2.45in]{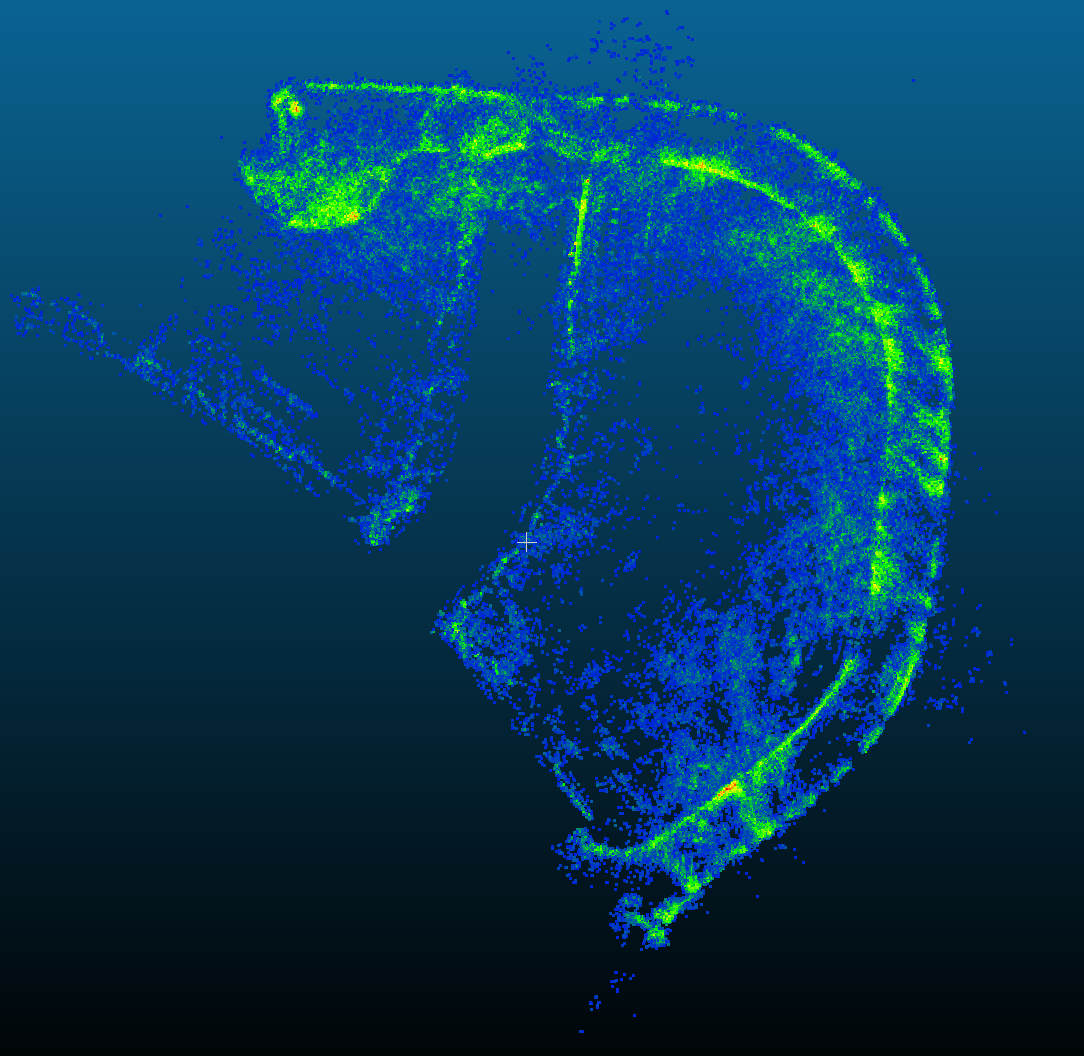}%
  \label{fig:binned_pointclouds}%
}
\caption{Example of binning raw radar detections into spatial histograms to elucidate the structure of the environment.}
\end{figure}

The principal challenge in creating useful maps from radar data is in distinguishing spurious detections from those that represent real features of the environment. Spurious detections are very common and they arise from many sources including multipath reflections, reflections from dynamic objects, reflections from the ground, ringing from signal processing, and others. This results in a very low repeatability between radar scans. This is demonstrated in Figure \ref{fig:raw_detections}, which shows the raw detections logged over the course of a short mission transformed to a common reference frame. The image shows very little identifiable structure in the aligned detections, which seem to be distributed evenly over the whole environment. 

An important characteristic of radar detections is not apparent from Figure \ref{fig:raw_detections}, however. As the robot moves through the environment true detections tend to appear in the same places while spurious detections tend to be spatially sparse. This means the structure of the environment becomes more apparent if the detections are binned into a spatial histogram where each voxel stores the number of detections that occur within it. Figure \ref{fig:binned_pointclouds} shows one such histogram colored by the number of detections per voxel, where detections are summed into the spatial histogram at the frequency of the sensor. One can see the structure of the environment is much more apparent in that image, with curved sidewalks and some primitive geometry of the neighborhood becoming clearer.

\begin{figure}[t]
\centering
\subfloat[A small subsection of radar OGM with observable features annotated.]{%
  \includegraphics[height=2in]{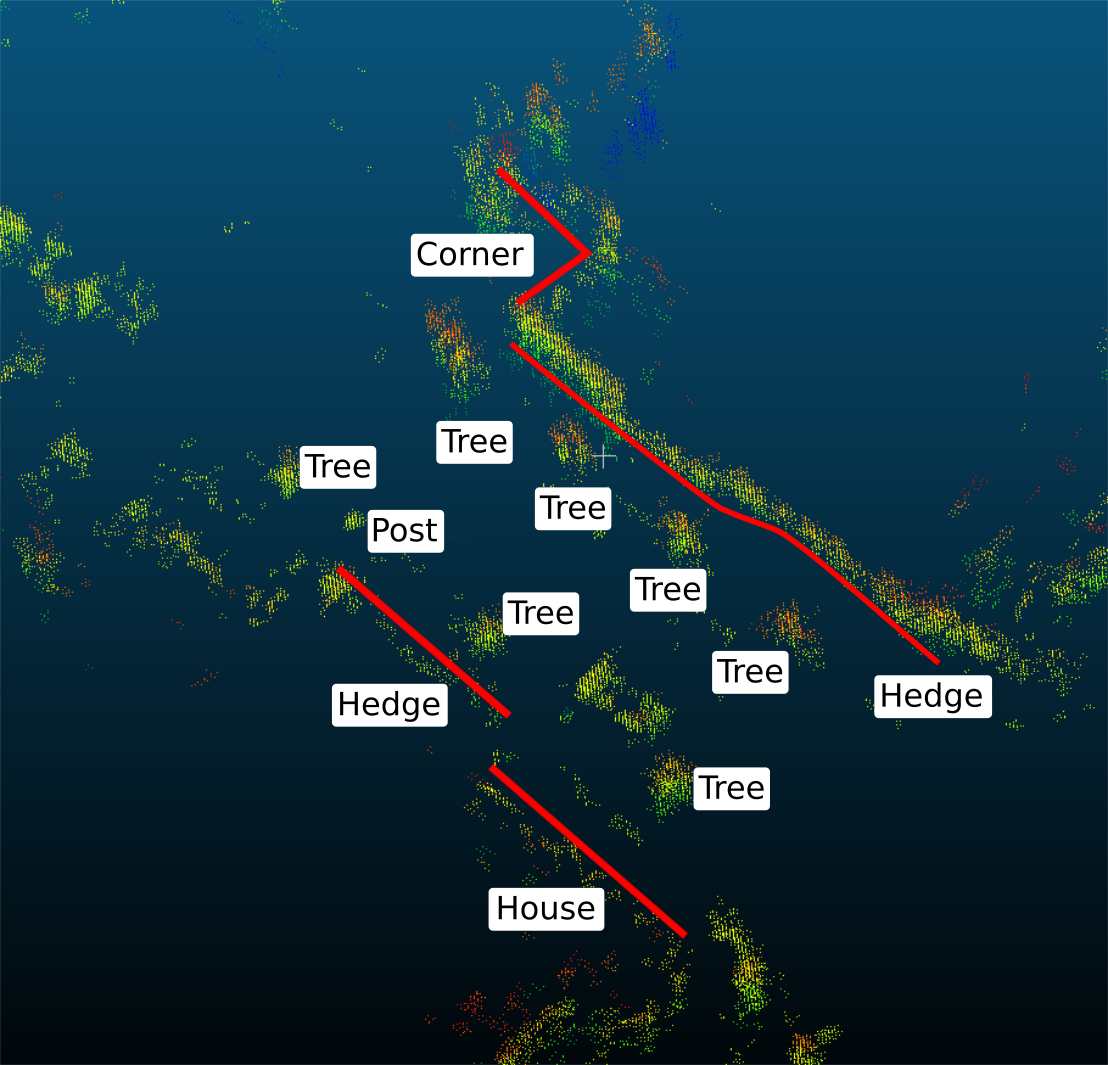}%
  \label{fig:ogm_detail}%
}\qquad
\subfloat[An aerial view of the same area represented in Figure \ref{fig:ogm_detail} with corresponding features annotated.]{%
  \includegraphics[height=2in]{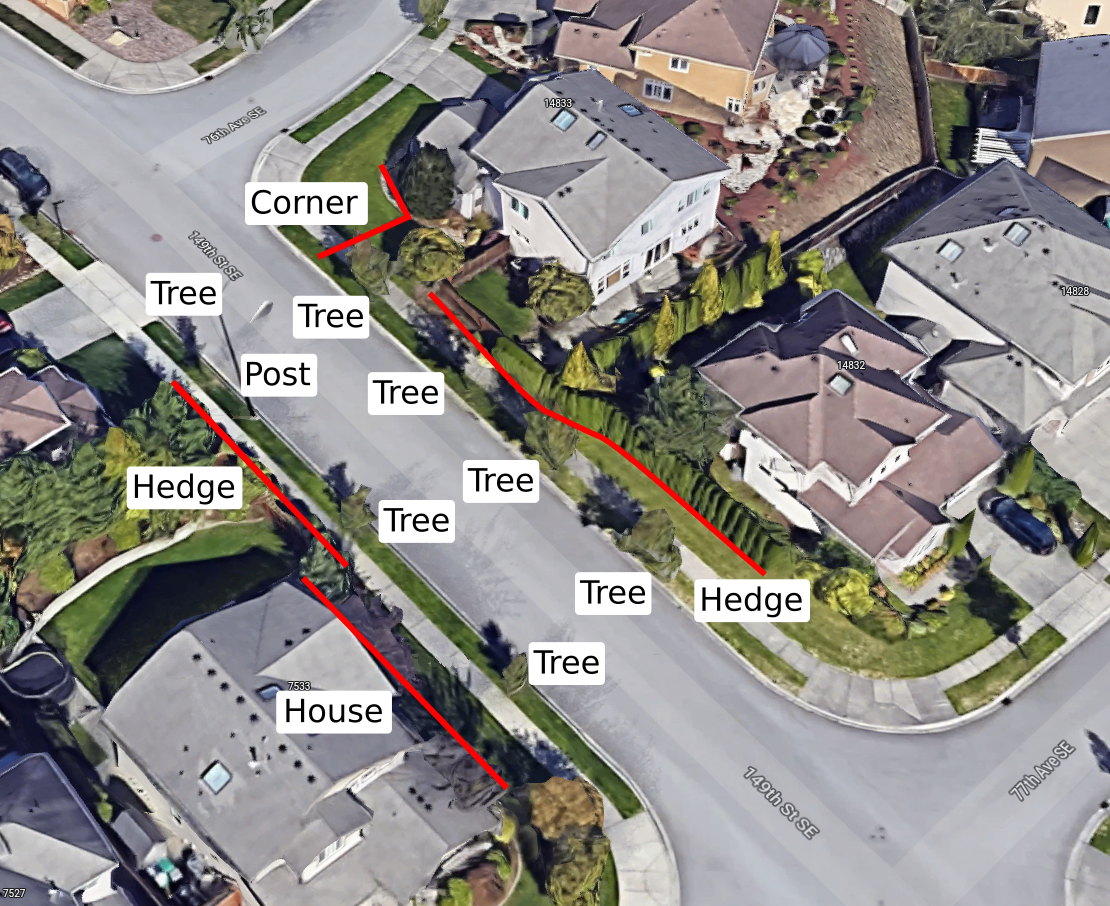}%
  \label{fig:aerial_map_detail}%
}
\caption{Detailed view of a radar-based OGM showing correspondences to an aerial view of the same area.}
\label{fig:map_detail}
\end{figure}

We take advantage of this property in our mapping method by first adding new radar detections to an occupancy grid map (OGM), which is a voxelized grid of 3D space parametrized by a probability from 0 to 1 which represents the probability that a specific voxel is occupied \cite{hornung2013octomap}. The overall OGM then has a tunable set of parameters, namely the likelihood threshold to consider a certain voxel ``occupied,'' and the probability of occupancy $P(n | z_t)$ where $n$ is a given voxel and $z_t$ is the current measurement, either a detection or non-detection. A probabilistic fusion is then applied using the posterior $P(n | z_{1:t-1})$ and an environmental prior $P(n)$ which is commonly assumed to be 0.5 (a uniform prior probability); our approach to this fusion is identical to the method of \cite{moravec1985high,hornung2013octomap}. We have found that through this method we can generate richly detailed radar maps that we have observed to remove map errors generally resulting from multipath reflections, ground reflections, and other radar-specific noise, while retaining the structure of the environment. An example OGM of a large area can be seen in Figure \ref{fig:maps}. Figure \ref{fig:map_detail} shows detail on a specific area of a radar OGM along with an aerial view of the same area to show the correspondence between features in the map and the real world.

Estimating the robot's pose involves registering a small local map of the robot's immediate surroundings, referred to as a query map, to a much larger map covering the whole area in which the robot is operating, referred to as a global map. The requirements for these two types of maps are different and the process for creating them differs as well. Creating a global map requires very high metric accuracy and global consistency over an area spanning whatever may be reasonably loaded onto the robot. In many applications, memory and runtime limitations drive this requirement, which may mean the map covers an area of 1--10 km$^2$. However, such a large map might contain millions of points, resulting in high memory overhead at runtime. Therefore, we further subdivide the neighborhood maps into chunks, which tessellate the global map and cover a smaller area, approximately 250m$^2$. The area to be mapped is traversed by a special mapping robot as shown in Figure \ref{fig:maps}. The radar sensor arrangement on the mapping robot is identical to that on the production robots, but the mapping robot's state estimation may differ; it may for instance be equipped with a high-accuracy GPS+INS system whose outputs are batch-optimized offline, or it may have a full data stream over a mission from which a fully bundle-adjusted trajectory may be resolved. In either case, as in many autonomous vehicle operations where prior maps are used for localization, the trajectory of the mapping robot must be centimeter-accurate in a global frame. Over the course of a mapping robot mission the detections from the robot's three radar sensors are also logged. After the mission the optimized trajectory is used to transform the logged detections to a common frame of reference. These transformed detections are then consumed into an OGM; the global map is represented as the point centers of the occupied cells of the OGM.

Query maps on the other hand must rely on only information available at hand to construct the OGMs, namely robot odometry, and therefore are subject to drift over time. Relative pose estimates from the RIO method described in \ref{sec:rio} have sufficient accuracy for transforming radar detections to the frame of the query map over a limited scale. Since many automotive-grade radars have accuracy on the order of 10cm at 30m range, query maps are only collected over an axis-aligned bounding box of $x, y \in $ 40m $\times $40m centered at the robot's current position. This clipping of data accomplishes two goals: first it limits the amount of odometric error that can infiltrate map generation, and second it bounds map size, since all voxels in the query map outside of this bounding box are cleared from memory. Note that this full node-traversal is required regardless, as the full data structure containing the map must be traversed in order to populate a point cloud for later consumption by map matching.

\subsection{Radar Map Matching}
\label{sec:map_matching}

The process for registering a query map to a global map proceeds in two stages during a mission. First, a ``Full'' comparison is performed, where coarse alignment is performed via a discretized brute-force search of possible alignments between the query and global maps. Then the best candidate alignment is refined using a smaller fine alignment, which also relies on brute-force searching, but terminates using the iterative closest point (ICP) method \cite{zhang_1992}. Second, once the ``Full'' search has concluded successfully, subsequent matching for later query maps may be performed using a ``Tracking'' comparison, where only the fine alignment and ICP steps are invoked. Since Tracking relies on a smaller number of comparisons and over a much smaller spatial scale, it concludes much more rapidly than the Full comparison, which generally is not required unless Tracking results in a failure to find a match.

During the coarse alignment phase, a discretized space of possible translations and rotations between the query map and global map are evaluated via brute-force search. For each possible alignment the points in the query map $\mathbf{p}_q$ are associated to their nearest neighbor in the global map $\mathbf{p}_g$ to create the set of matched points $\mathcal{P}$. A score $s$ for each alignment is calculated as the the number of points in the query map that are within $d$ of the global map:

\begin{equation}
\label{eq:score}
    s = \sum_{\textbf{p}_q,\textbf{p}_g \in \mathcal{P}}
    \begin{cases}
        1, & \text{if } \|\textbf{p}_q - \textbf{p}_g\| < d \\
        0, & \text{otherwise }
    \end{cases}
\end{equation}

\noindent where $d$ is a threshold for a point to be an inlier.

Spatial pyramiding \cite{olson2015m3rsm} is employed to make the brute-force search process more impervious to defective matches resulting from local minima of the cost surface induced by Eq.\ \eqref{eq:score}. The top $K$ scoring alignments from pyramid level $n$ are refined in pyramid level $n+1$ and the process is repeated until the highest-resolution pyramid level is reached. \new{In our methodology we search within $\pm \pi/2$ in yaw and $\pm 5m$ in translation for the ``Full'' search, and in ``Tracking'' limit our search to $\pm \pi/4$ and $\pm 1m$ which we find to yield few losses of tracking, generally conditioned on the environment and the presence of map features to be matched.}

The highest scoring alignment from the brute-force search process are then refined using ICP with the standard point-to-point metric. A score is calculated for the refined alignment using equation \ref{eq:score}. If this score is equal to or above a threshold value, the alignment is considered to have been successful.

\section{Experiments}

\begin{figure}[h!]
\centering
\includegraphics[width=0.45\linewidth]{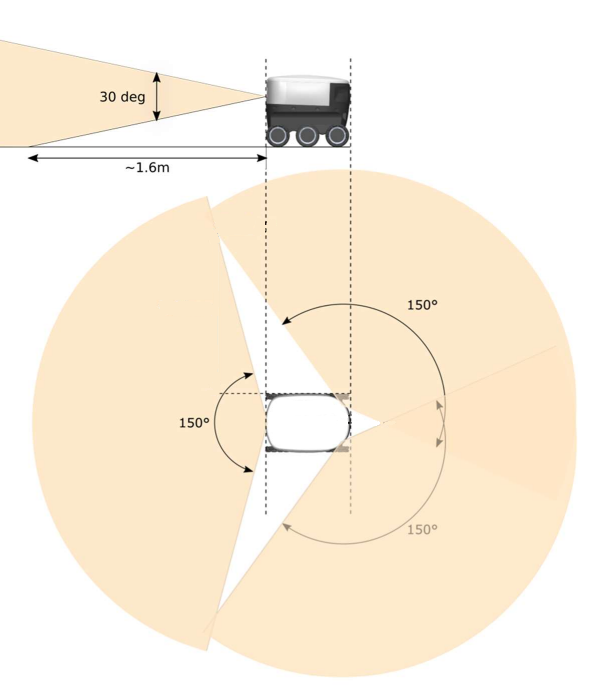}
\caption{Fields of view of the robot's radar sensors. Above a side view of the robot shows the rear sensor's vertical field of view and below a top-down view shows all three sensors' horizontal fields of view.}
\label{fig:radar_fov}
\end{figure}

We conducted a series of experiments evaluating the accuracy, reliability, and computational requirements of our odometry and relocalization methods. Both experiments used the same autonomous ground vehicle which was equipped with 3 automotive radars, a Lord Microstrain 3DM-GX5 IMU, and a high-accuracy Novatel GPS+INS system for groundtruth. The vehicle and sensor locations are detailed in Figure \ref{fig:radar_fov} and specifications of the radar sensors are detailed in Table \ref{tab:radar_specs}. \new{We acknowledge the existence of some excellent datasets for radar \cite{choi2023msc,kramer2021coloradar,barnes2020oxford}, each relying on their own radar sensor(s) to encourage the development of novel capabilities. In our case, our radar was a proprietary blend of hardware and signal processing techniques, limiting the capability of our testing on different datasets.}

\begin{table}
\centering
 \begin{tabular}{| c | c |} 
\hline
 max range & $80\text{m}$ \\ [0.5ex]
 \hline
 range resolution & $0.3\text{m}$ \\ [0.5ex]
 \hline
 range accuracy & $0.03\text{m}$ \\  [0.5ex] 
 \hline
 Doppler min/max & $\pm30\text{m/s}$ \\ [0.5ex]
 \hline
 Doppler resolution & $0.4\text{m/s}$ \\ [0.5ex]
 \hline
 Doppler accuracy & $0.04\text{m/s}$ \\ [0.5ex]
 \hline
 azimuth FOV & $\pm75^\circ$ \\ [0.5ex]
 \hline
 azimuth resolution & $2^\circ$ \\ [0.5ex]
 \hline
 azimuth accuracy & $0.2^\circ$ \\ [0.5ex]
 \hline
 elevation FOV & $\pm15^\circ$ \\ [0.5ex]
 \hline
 elevation resolution & $2.5^\circ$ \\ [0.5ex]
 \hline
 elevation accuracy & $0.25^\circ$ \\ [0.5ex]
 \hline
 Update Rate & $20$Hz \\ [0.5ex]
 \hline
 Frequency & $76-77$GHz \\ [0.5ex]
 \hline
\end{tabular}
\vspace{1mm}
\caption{Specifications of the radar sensors used in field experiments.} \label{tab:radar_specs}
\end{table}

\subsection{Radar-Inertial Odometry Evaluation}

To evaluate the RIO method we use four missions carried out in residential neighborhoods. Collectively, the missions cover approximately 12km in length and 4 hours in duration. We report statistics on the error accumulated by our RIO method over non-overlapping 10m subsegments of each trajectory using the GPS+INS system for groundtruth. We report the median (p50), 95th percentile (p95), 99th percentile (p99), and maximum drift for both translation and heading for each mission in Table \ref{tab:odom_errors}.

\begin{table}
\centering
 \begin{tabular}{|c | c | c | c |} 
 \hline
 \multicolumn{2}{|c|}{} & \thead{Translation Drift\\(m/m)} & \thead{Heading Drift\\($^\circ/\text{m}$)} \\
 \hline
 \multirow{4}{*}{\thead{Mission 1}}
 & p50 & 0.013 & 0.021 \\
 & p95 &  0.027 & 0.084 \\
 & p99 & 0.052 & 0.254 \\
 & max & 0.275 & 0.483 \\
 \hline
 \multirow{4}{*}{\thead{Mission 2}}
 & p50 & 0.017 & 0.037 \\
 & p95 &  0.043 & 0.123 \\
 & p99 & 0.064 & 0.194 \\
 & max & 0.127 & 0.224 \\
 \hline
 \multirow{4}{*}{\thead{Mission 3}}
 & p50 & 0.015 & 0.033 \\
 & p95 &  0.035 & 0.105 \\
 & p99 & 0.042 & 0.162 \\
 & max & 0.078 & 0.99 \\
 \hline
 \multirow{4}{*}{\thead{Mission 4}}
 & p50 & 0.017 & 0.034 \\
 & p95 &  0.041 & 0.119 \\
 & p99 & 0.050 & 0.176 \\
 & max & 0.072 & 0.248 \\
 \hline
\end{tabular}
\vspace{1mm}
\caption{Translation and heading drift statistics for each mission.} \label{tab:odom_errors}
\end{table}

With median translational drift as low as $0.013$m/m and heading drift as low as $0.021^\circ/\text{m}$, our RIO method is impressively accurate. It is very lightweight as well. When running on an Intel Core i7-10870H processor our RIO method's mean update time is just 0.86ms. On an ARM Cortex-A57 (a common low-power embedded processor) its mean update time is still only 4ms. Our radar sensor's update period is 50ms, so even on low-power embedded hardware our RIO method leaves plenty of headroom for other processes.

\subsubsection{Ablation Study}

To evaluate the importance of different components of the proposed RIO system we re-ran the system offline with different components de-activated. To evaluate the benefit of using multiple synchronized radar sensors we include an experiment in which only the front-left sensor is used. To evaluate the benefit of the heading constraint described in Section \ref{sec:heading_constraint} we include an experiment in which the heading constraint is not used. Table \ref{tab:ablation} shows the results of these experiments. 

\begin{table}
\centering
 \begin{tabular}{| c | c | c | c |} 
 \hline
 \multicolumn{2}{|c|}{} & \thead{Translation Drift\\(m/m)} & \thead{Heading Drift\\($^\circ/\text{m}$)} \\
 \hline
 \multirow{4}{*}{\thead{Three Radar}} 
 & p50 & 0.017 & 0.034 \\
 & p95 &  0.041 & 0.119 \\
 & p99 & 0.050 & 0.176 \\
 & max & 0.072 & 0.248 \\
 \hline
 \multirow{4}{*}{\thead{Single Radar}}
 & p50 & 0.017 & 0.059 \\
 & p95 &  0.070 & 0.195 \\
 & p99 & 0.141 & 0.309 \\
 & max & 0.314 & 0.493 \\ 
 \hline
 \multirow{4}{*}{\thead{No Heading\\Constraint}} 
 & p50 & 0.049 & 0.526 \\
 & p95 &  0.257 & 2.258 \\
 & p99 & 0.322 & 4.714 \\
 & max & 0.392 & 15.59 \\
 \hline
\end{tabular}
\vspace{1mm}
\caption{Translation and heading drift statistics for mission 4 with different components of the odometry system de-activated.} \label{tab:ablation}
\end{table}

We can see from Table \ref{tab:ablation} that using only one radar still performs reasonably well. The median and even the 95th percentile drift statistics for the single radar version are fairly close to the three radar version. The benefit to using multiple radar sensors seems to be in the tails of the error distribution because the 99th percentile and maximum drift statistics for the single radar version are far higher than the three radar version. The version with the heading constraint de-activated does not perform nearly as well, however. The translation drift statistics for the no-heading-constraint version are 3 to 5 times those for the three radar version and the heading drift statistics are 16 to 60 times higher. Clearly the heading constraint is crucial for accurate odometry estimation.

\subsection{Radar Map Matching Evaluation}

To evaluate our radar mapping and matching method we created a global map of a neighborhood shown in Figure \ref{fig:maps} using our mapping robot. Then the robot traverses the same path twice continuously creating local query maps and attempting to register them to the global map. The mapping robot's Novatel GPS+INS system is used both to create the global map, and to evaluate the accuracy of alignments between the query and global maps. The global map covers a square area roughly 220m on a side, the query maps are roughly 20m on a side, and both the global and query maps use a 0.2m resolution in their OGMs. Within this global map region, the robot traverses a trajectory of 1.6km. 

We evaluate the quality of our map matching method by comparing the transforms calculated from successful map registrations to groundtruth transforms from the Novatel GPS+INS system. We also note the availability of map matches as the number of successful matches over the total number of attempted matches. We compare the results for radar map matching to a our own lidar map matching method, which uses LOAM-generated submaps \cite{zhang2014loam} as the query map (rather than an OGM) but otherwise uses the same coarse-to-fine Full and Tracking comparison methods as our radar map matching method. 

Table \ref{tab:map_matching} shows the errors in translation estimates for radar and lidar map matching for both missions. One can see the median errors for radar and lidar map matching are similar, although lidar map matching gives lower median error. Radar map matching clearly outstrips lidar at the tails of the error distribution though. Radar's 90th and 99th percentile errors are far lower than those for lidar. 

\begin{table}
\centering
 \begin{tabular}{| c | c | c | c | c | c |} 
 \hline
 \multicolumn{2}{|c|}{} & \multicolumn{2}{|c|}{\thead{Radar Translation \\ Error (m)}} & \multicolumn{2}{|c|}{\thead{Lidar Translation \\ Error (m)}} \\
 \hline
 \multicolumn{2}{|c|}{} & $x$ & $y$ & $x$ & $y$ \\
 \hline
 \multirow{4}{*}{\thead{Mission 1}} 
 & p50 & 0.1757 & 0.199 & 0.1791 & 0.2111 \\
 & p90 & 0.5056 & 0.4903  & 1.8316 & 1.3199\\
 & p99 & 0.7995 & 0.9167  & 8.324 & 4.9173\\
 & max & 1.6315 & 2.9356  & 10.5059 & 6.8023\\
 \hline
 \multirow{4}{*}{\thead{Mission 2}}
 & p50 & 0.2205 & 0.2161 & 0.1675 & 0.1932 \\
 & p90 &  0.6081 & 0.54  & 1.3718 & 1.932\\
 & p99 & 0.8813 & 0.7492  & 4.6913 & 6.5304\\
 & max & 1.0016 & 0.8195  & 5.808 & 7.5731\\
 \hline
\end{tabular}
\vspace{1mm}
\caption{Translation error for radar and lidar map matching techniques.} \label{tab:map_matching}
\end{table}

\begin{figure}[t]
\centering
\includegraphics[width=0.45\linewidth]{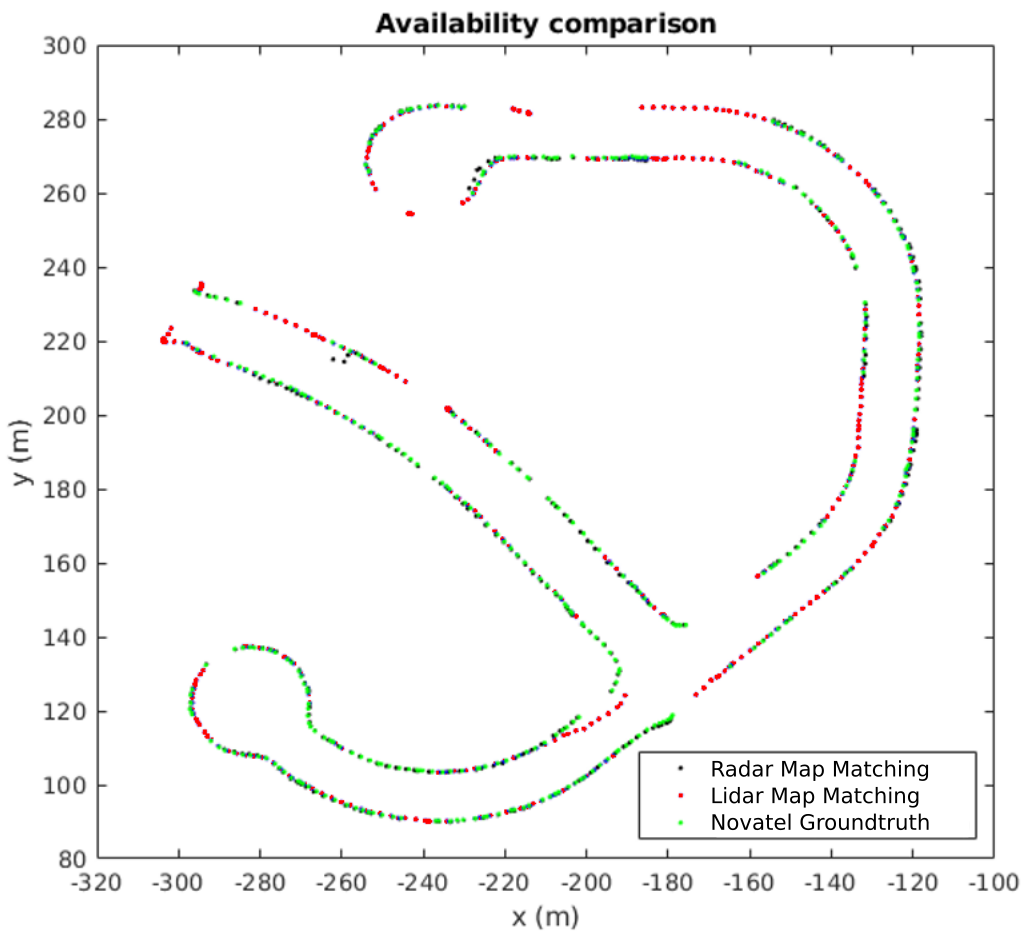}
\caption{Comparison of the poses estimated by our radar map matching method with lidar map matching with radar pose estimates in black, lidar pose estimates in red, and groundtruth poses for each estimated radar pose in green.}
\label{fig:availability}
\end{figure}

The advantage of radar map matching in accuracy is somewhat compensated by its lower availability, however. Out of 1715 attempted matches, our radar map matching method succeeds 1389 times for a total availability of 0.81. Lidar, on the other hand matches 1652 out of 1826 for a total availability of 0.90. This is demonstrated in Figure \ref{fig:availability}, which shows the poses estimated by lidar map matching in red, radar map matching in black, and a groundtruth pose for each radar pose in green. One can see from this image that lidar map matching successfully finds matches more often, but its pose estimates can also be less accurate.

\section{Conclusion} 
\label{sec:conclusion}

In this paper we presented novel radar-based methods for odometry and global localization. For our radar-inertial odometry (RIO) method we combine an established Doppler velocity constraint with a novel method for dynamic outlier rejection and a unique heading constraint based on radar landmarks. We demonstrate the combination of these elements results in unprecedented relative pose estimation accuracy for a method using automotive sensors. Our RIO method's accuracy is comparable to that of some methods based on high-resolution NavTech sensors. For example, our RIO method achieves median translation drift of 1.7\% in the worst case while \textit{Under The Radar} achieves 2.1\% \cite{barnes_2020} and \textit{Masking By Moving} achieves 1.6\% \cite{barnes_2019} \new{according to their evaluation. We note however that a direct comparison on their data is not possible given the unique experimental platform for which our method was developed}. Further, our proposed RIO method is shown to be impressively lightweight; it can be used on low-power embedded hardware with ample headroom for other processes. 

Our radar-based mapping method is able to aggregate radar detections into a representation of the robot's environment that is repeatable and highly descriptive. This is demonstrated by our method for registering radar maps, which is able to achieve relative pose estimation accuracy similar to lidar-based methods. Unlike lidar, radar is sensitive to the properties of materials in addition to their geometry. This means radar maps can represent changes in material properties where there is no corresponding change in geometry as in the cases of access hole covers, storm drain grates, etc. This makes radar map registration less susceptible to geometric ambiguity, as demonstrated by the much lower 99th percentile and maximum pose estimation errors of radar map matching relative to lidar. \new{Our primary finding is that by balancing the IMU, Doppler, and landmark factors in RIO, as well as utilizing an incrementally-built OGM for map matching, we are able to achieve localization performance acceptable for suburban neighborhood navigation.}

Together, our RIO and radar map matching methods provide a highly robust localization system for autonomous ground vehicles (AGVs) using only radar and inertial measurements. Radar map matching is able to globally localize the vehicle with respect to a pre-built map, and RIO can provide smooth and continuous pose updates between map matches. These developments could allow for the operation of AGVs in far more diverse and challenging environments from complete darkness to rain, snow, smoke and fog.

\section*{Acknowledgments}
This work was done for and supported by Amazon's Scout project.

\bibliographystyle{apalike}
\bibliography{references}

\end{document}